\documentclass{article}

\usepackage{microtype}
\usepackage{graphicx}
\usepackage{caption}
\usepackage{subcaption}
\usepackage{booktabs} 
\usepackage{tabularx}
\usepackage{enumitem}
\usepackage{hyperref}

\usepackage[accepted]{icml2023}

\usepackage{amsmath}
\usepackage{amssymb}
\usepackage{mathtools}
\usepackage{amsthm}
\usepackage{multicol}

\usepackage[capitalize,noabbrev]{cleveref}

\theoremstyle{plain}

\theoremstyle{definition}

\theoremstyle{remark}

\usepackage[textsize=tiny]{todonotes}

\icmltitlerunning{The Synergy of Speculative Decoding and Batching in Serving LLMs}

\begin{document}

\twocolumn[
\icmltitle{The Synergy of Speculative Decoding and Batching\\ in Serving Large Language Models}


\icmlsetsymbol{equal}{*}

\begin{icmlauthorlist}
\icmlauthor{Qidong Su\footnotemark[1]\footnotemark[2]\footnotemark[3]}{}
\icmlauthor{Christina Giannoula\footnotemark[1]}{}
\icmlauthor{Gennady Pekhimenko\footnotemark[1]\footnotemark[2]\footnotemark[3]}{}
\end{icmlauthorlist}

\icmlkeywords{Machine Learning, ICML}

\vskip 0.3in
]

\footnotetext[1]{University of Toronto}
\footnotetext[2]{CentML Inc}
\footnotetext[3]{Vector Institute}




\begin{abstract}

Large Language Models (LLMs) like GPT are state-of-the-art text generation models that provide significant assistance in daily routines. However, LLM execution is inherently sequential, since they only produce one token at a time, thus incurring low hardware utilization on modern GPUs. Batching and speculative decoding are two techniques to improve the GPU hardware utilization in LLM inference. To study their synergy, we implement a prototype implementation and perform an extensive characterization analysis on various LLM models and GPU architectures. We observe that the optimal speculation length depends on the batch size used. We analyze the key observation and build a quantitative model to explain it. Based on our analysis, we propose a new adaptive speculative decoding strategy that chooses the optimal speculation length for different batch sizes. Our evaluations show that our proposed method can achieve equal or better performance than the state-of-the-art speculation decoding schemes with fixed speculation length.
\end{abstract}

    

\section{Introduction}

Large Language Models (LLMs) such as GPT-3~\cite{gpt3}, GPT-4~\cite{gpt4}, and LLama~\cite{llama} have achieved great success in not only natural language processing, but also other scientific and technical fields, including robotics~\cite{brohan2023rt}, programming languages~\cite{llm4code}, and medical science~\cite{thirunavukarasu2023large}. 

However, text generation, the most common usage of LLMs, is carried out in a sequential execution paradigm called auto-regression. In auto-regression, LLMs predict the subsequent \emph{one} token based on previous tokens. The newly generated token will be appended to the end of the sequence of generated tokens. In each iteration, the input query tensor only contains the embedding of one token (except for the first iteration), and only one token will be generated. This leads to a low amount of execution parallelism and GPU hardware utilization. To make things worse, the execution of LLM inference is usually memory-bounded~\cite{ivanov2021data}. The sequential execution paradigm requires GPUs to load the huge weight matrices from off-chip memory to on-chip memory in each iteration, which leads to poor memory reuse and performance.

Researchers have explored two approaches to tackle the aforementioned problem: (i) batching~\cite{orca, fang2021turbotransformers}, and (ii) speculative decoding~\cite{specsample, fastinferspec, specinfer, spector2023accelerating, liu2023online}. First, batching, as used in other deep-learning applications~\cite{gao2018low, crankshaw2017clipper, lee2018pretzel, olston2017tensorflow, shen2019nexus}, merges several user requests together as one batched request, thus GPUs produce one new token in parallel for each user request in one batch. Second, speculative decoding uses a smaller model (Small Speculative Model, SSM) to predict more than one subsequent token (henceforth referred to as \emph{speculation length}) and uses the original LLM to verify the prediction in parallel. Verifying a sequence of tokens is much faster than sequentially generating them thanks to the higher GPU hardware utilization. Although the SSM also provides subsequent tokens in a sequential manner, its overhead is much smaller than the runtime of LLM, thus it is amortized. Therefore, if the SSM can accurately mimic the behavior of the original LLM, the end-to-end runtime of inference can be significantly reduced.

Our goal in this work is to study the synergy of these two approaches and how we can get the best of these two approaches. Intuitively, they are both designed to saturate the GPU hardware resource, 
enabling one of them might offset the benefits of the other one. Enabling both of them simultaneously might be ineffective or even worse. Also, batching is not always available as it is constrained by the request traffic volume and available GPU memory size. 

To this end, we provide an extensive characterization study of batching and speculative decoding execution. First, we carried out a comprehensive profiling study of batched speculative decoding using different models, GPUs, batch sizes, and speculation lengths. 
  
Our key observation is that the optimal speculation length varies across different batch sizes. More specifically, larger batch sizes require a smaller speculation length to achieve optimal performance, and a speculation length too large will deteriorate the performance. We developed an analytical model to explain this phenomenon. This necessitates the design of an adaptive strategy that finds the optimal configuration for different execution scenarios.

Therefore, we propose an adaptive speculative decoding strategy, which adjusts the speculation length according to the batch size used. It runs a short period of profiling before deployment and builds a mapping from the batch size to its corresponding optimal speculation length. After the service is launched, it will use the optimal speculation length for an arriving batch of requests.

Our evaluations show that our proposed method achieves equal or better performance than the state-of-the-art speculation decoding schemes with fixed speculation length. For time-varying requests, adaptive speculative decoding can provide an extra 9\% latency reduction compared with fixed speculation length.

In conclusion, our major contributions are:
\begin{itemize}[noitemsep,topsep=0pt,leftmargin=8pt]
\item We carry out a comprehensive study to investigate how the batch size interacts with the speculation length. We observe that the optimal speculation length depends on the batch size: smaller speculation lengths need to be chosen for large batch sizes.
\item We analyze the observations of our study by building a quantitative model to explain them.
\item We propose a new adaptive speculative decoding strategy that chooses the optimal speculation length for different batch sizes at low latency costs.
\item We evaluate the proposed strategy and compare it with prior schemes with fixed speculation length. Our experiments show that our proposed method can achieve equal or better performance than the prior schemes.
\end{itemize}
\section{Background}



\noindent\textbf{Batching.} Batching~\cite{orca} is one of the most popular performance optimizations in deep learning. It merges multiple data instances into one, to increase parallelism and hardware utilization. It can increase the throughput, especially when the batch size is small.  In the context of LLMs, multiple user requests that arrive at roughly the same time can be processed in a batch. However, batching also demands more memory usage since the intermediate result of all requests in a batch needs to be stored. The effectiveness of batching is also bounded by the request traffic.


\noindent\textbf{Speculative Decoding.} Speculative decoding~\cite{specsample, fastinferspec, specinfer} is another method to increase execution parallelism. It uses a Small Speculative Model (SSM) to first predict several subsequent tokens, and feed them into the original Large Language Model (LLM), which verifies the predictions and corrects incorrect predictions. The verification for each token can be parallelized, thus increasing the hardware utilization. The process of speculation decoding is shown in Algorithm~\ref{alg:spec}. 

In each step, the SSM predicts $s$ subsequent tokens ($t_1, t_2, \dots, t_s$), which will be verified by the LLM later. Verification is done by calculating the logits corresponding to each token ($o_1, o_2, \dots, o_s$). The $i$-th logit $o_i$ is a probability distribution across the whole vocabulary, representing which token should be the $(i+1)$-th token, if all previous tokens are correct. If for some index $l$, $t_{l+1}$ is not the token with the highest probability in $o_i$, all tokens after $t_{l}$ will be discarded, i.e. the correctness of one speculated token relies on the correctness of its previous tokens. 

Assuming $l$ of them are correct, the length of the correct sequence increases by $l + 1$, since the LLM provides either a correction or an extra look-ahead token, which is inferred from the logits of the last correct token. One point worth noting is that even if the SSM never produces a correct prediction, the algorithm can still terminate since the LLM can always generate one new correct token in each iteration ($\mathrm{argmin}(o_l)$ in the algorithm above). 

\begin{algorithm}[h]
\caption{Speculative Decoding (argmax sampling)}
\label{alg:spec}
\begin{algorithmic}
\STATE {\bfseries Input:} Input tokens $I$, speculation length $s$
\STATE $S$ = $I$
\STATE Initialize KV cache $kv_{SSM}$, $kv_{LLM}$
\WHILE{true}
    \STATE $t_1, t_2, \dots, t_s = speculate(S, kv_{SSM}, s)$
    \STATE $o_1, o_2, \dots, o_s = LLM(S, t_1, t_2, \dots, t_s, kv_{LLM})$
    \STATE $l = $ first index that $t_l \neq \mathrm{argmin}(o_l)$
    \STATE $S = S \oplus t_1, t_2, \dots, t_{l-1}\oplus\mathrm{argmin}(o_l)$
    \IF{$\texttt{<EOS>} \in S$}
        \STATE \bfseries{break}
    \ENDIF
\ENDWHILE
\STATE \bfseries{return} $S$ 
\end{algorithmic}
\end{algorithm}

\section{Batched Speculative Decoding}

In this section, we describe our implementation of batched speculative decoding and show the benchmark result of if for different models, GPUs, batch sizes, and speculation lengths. We also include empirical and modeling analysis to explain this phenomenon.

\subsection{Prototype Implementation.} To investigate the synergy of speculative decoding and batching, we implemented a prototype implementation supporting both these optimizations for HuggingFace transformers models~\cite{huggingface}.
The correction of wrong speculation is implemented using attention masks: if one token does not pass the verification of the LLM, the current token and all subsequent tokens after it will be discarded by masking them off.
The correct token given by the LLM is then appended to the end of the sequence, which will be used in the next iteration. 


\subsection{Experimental Analysis}

\begin{figure}[!ht]
    \centering
    \begin{subfigure}[b]{0.23\textwidth}
        \centering
        \includegraphics[width=\textwidth]{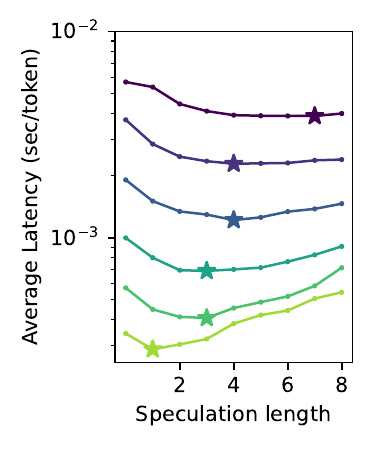}
        \caption{RTX 3090, OPT-1.3B}
        \label{fig:3090-opt13}
    \end{subfigure}
    \hfill
    \begin{subfigure}[b]{0.23\textwidth}
        \centering
        \includegraphics[width=\textwidth]{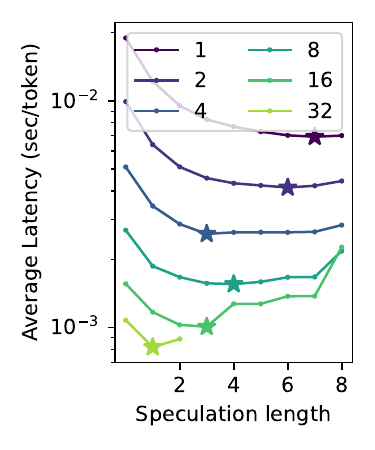}
        \caption{RTX 3090, OPT-6.7B}
        \label{fig:3090-opt67}
    \end{subfigure}
    \\
    \begin{subfigure}[b]{0.23\textwidth}
        \centering
        \includegraphics[width=\textwidth]{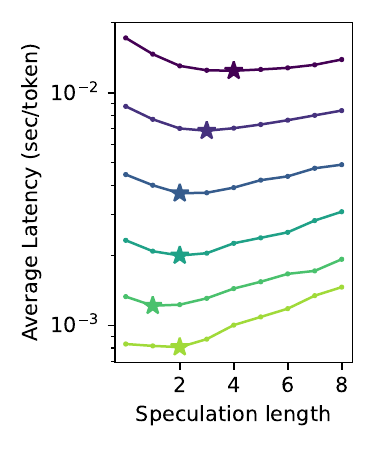}
        \caption{A100, OPT-6.7B}
        \label{fig:a100-6.7B}
    \end{subfigure}    
    \hfill
    \begin{subfigure}[b]{0.23\textwidth}
        \centering
        \includegraphics[width=\textwidth]{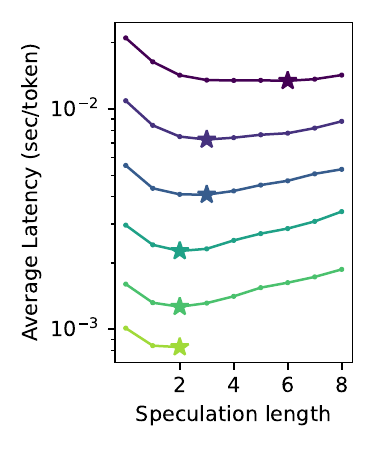}
        \caption{RTX 4090, OPT-6.7B}
        \label{fig:4090-6.7B}
    \end{subfigure}
    \\
    \begin{subfigure}[b]{0.23\textwidth}
        \centering
        \includegraphics[width=\textwidth]{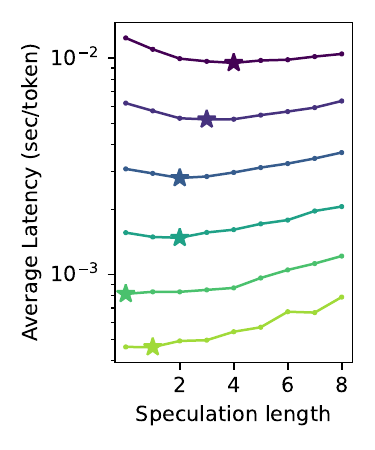}
        \caption{RTX 4090, OPT-1.3B}
        \label{fig:4090-1.3B}
    \end{subfigure}\hfill
    \begin{subfigure}[b]{0.23\textwidth}
        \centering
        \includegraphics[width=\textwidth]{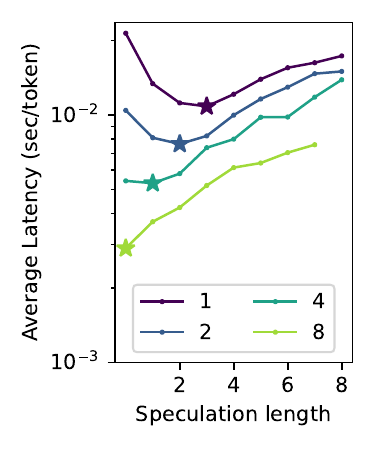}
        \caption{RTX 3090, Llama-7B}
        \label{fig:3090-llama}
    \end{subfigure}

    \caption{How per token latency changes with the batch size and speculation length, for different models and GPUs. The optimal speculation length is highlighted. A common phenomenon is that the optimal speculation length is smaller for larger batch sizes.}
    \label{fig:measure}
\end{figure}

We measure the average latency per token achieved for different combinations of batch sizes (from 1 to 32 with steps to be a power of 2) and speculation lengths (from 1 to 8) in various settings: (i) in Figure~\ref{fig:3090-opt13}, ~\ref{fig:3090-opt13}, ~\ref{fig:3090-opt67} 
we use RTX 3090 GPU model and vary the LLM model (OPT-1.3B, OPT-6.7B~\cite{opt} and Llama-7B~\cite{llama}), (ii) in Figure~\ref{fig:3090-opt67}, \ref{fig:4090-6.7B}, and \ref{fig:a100-6.7B} we use the OPT-6.7B model and we vary the GPU architecture (RTX 3090, RTX 4090, A100). In these figures, the asterisk annotated point corresponds to the optimal speculation point for each different batch size.


We make four key conclusions. First, combining batching with speculative decoding provides performance improvements. For example, in Figure~\ref{fig:3090-opt67} with batch size 4, having speculation length 3 improves speed by $1.93\times$. 
Second, we observe that the benefits of speculative decoding are high when using small batch sizes, i.e., increasing the speculation length improves the per-token latency in smaller batch sizes. 
When the batch size is 1, speculative decoding reduces the per-token latency by up to 63\%.  This is because GPU utilization is low with small batch size thus using higher speculation length helps.
Third, we find that when using large batch sizes (e.g., 16 or 32), higher speculation lengths incur performance slowdowns. When the batch size is 32, the smallest per-token latency is achieved using a speculation length smaller or equal to 2. This is because large batch sizes already fully utilize the underlying GPU computational resources, thus improving the execution parallelism with speculative decoding cannot provide further benefits. Fourth, the best-performing speculation length depends on the batch size and the characteristics of GPU architecture and LLM model. With a batch size of 8 in RTX3090, the optimal speculation length is  3 for OPT-1.3B and 4 for OPT-6.7B model. Similarly, using batch size 4 in OPT-6.7B execution, the optimal speculation length is 2 for A100 and 3 for RTX 4090. 
We conclude that an adaptive policy is necessary to autotune the speculation length based on the bathe size and the characteristics of LLM model and GPU architecture.




\subsection{Modeling Analysis}\label{sec:model}

To analyze the reason behind the aforementioned LLM behavior, we model the total execution runtime of batched LLM inference with speculative decoding. Table~\ref{tab:notation} lists the notations that we use in our modeling. 


Intuitively, the total execution runtime with speculative decoding consists of two parts: (i) the runtime of LLM (\textbf{$T_L$}), and (ii) the runtime of the SSM (\textbf{$T_S$}).

Both $T_L$ and $T_S$ depend on the batch size $b$ and the speculation length $s$. Thus, we formulate the total execution runtime as follows:
\begin{equation}
T_{total} = T_L(b, s) + T_S(b, s)
\end{equation}

For simplicity, we only consider the runtime of the auto-regression stage, neglecting the prefilling stage (i.e. the first iteration where the SSM and LLM reads the whole prompt). 
We also adopt the assumption that each iteration of these two models takes the same time. This is true when the context is not extremely long, and the bulk of the runtime is spent on matrix multiplications other than attention, which is proportional to the input query length. During the auto-regression stage, the length of the input query is fixed.

Based on these assumptions, we can write the total runtime of the LLM and the SSM as the product of the number of calls to them and the runtime of each call as follows. Used notations are listed in Table~\ref{tab:notation}. As mentioned in Algorithm~\ref{alg:spec}, in each iteration, the LLM always provides one extra correct token than $l(s)$. Therefore, the number of iterations is $\frac{N}{l(s)+1}$. We can thus write the runtime of the LLM and SSM as:
\begin{align}
&T_L(b,s) =  \frac{N }{l(s) + 1} \cdot t_L(b, s)\\
&T_S(b,s) =  \frac{N }{l(s) + 1} \cdot \left( s \cdot t_S(b, 1) \right)
\end{align}

\begin{table}[t]
    \centering
    \small
    \caption{Notations used for modeling the LLM execution runtime.}
    \begin{tabular}{ p{0.2\linewidth} | p{0.7\linewidth} }
        \toprule
        $N$ & The total number of tokens to be generated \\ \midrule
        $b$ & The batch size to be used in inference \\ \midrule
        $s$ & The number of speculated tokens generated at each speculative step \\ \midrule
        $l(s)$ & Average number of correct tokens generated by the SSM\\ \midrule
       $T_{L}(b, s)$ & Total execution time of the LLM \\ \midrule
       $T_{S}(b, s)$ & Total execution time of the SSM \\ \midrule
       $t_{L}(b, s)$ & Per step runtime of the LLM with batchsize=$b$ and query length $s$ \\ \midrule
       $t_{S}(b, s)$ & Runtime of the SSM with batchsize=$b$ and query length $s$ \\
       \bottomrule
    \end{tabular}
    \label{tab:notation}
\end{table}

\subsubsection{LLM Runtime $T_L$} 


LLM is used to verify the tokens generated by the SSM, and generates a new token to either correct wrong speculations or look one token ahead if the speculation is correct.
Assuming $l(s)$ provides the average number of correct tokens generated by SSM, the total number of tokens generated in the iteration is ${l(s) + 1}$. 
Thus, given that $N$ is the number of token to be generated and $t_{L}(B, s)$ is the execution time for LLM to generate a token, the total LLM runtime can be estimated as  $\frac{N}{l(s) + 1} \cdot t_{L}(B, s)$.


\noindent\textbf{Expected Number of Correct Tokens $l(s)$}

\begin{figure}
    \centering
    \includegraphics[width=0.8\linewidth]{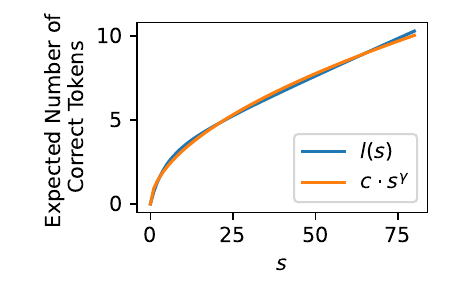}
    \caption{How $l(s)$ scales with $s$, which can be approximated by a sublinear power function $c\cdot s^\gamma$, where $\gamma < 1$. The orange curve is an approximation 
 $0.9s^{0.548}$.
    }
    \label{fig:expected-len}
\end{figure}

The expected number of correct tokens $l(s)$ represents how accurately the SSM can mimic the behavior of the LLM. While it is hard to directly model it with a closed form, we did an empirical measurement to observe how it scales with $s$.
We sampled $n=200$ prompts as inputs and let the SSM generate $m=80$ tokens for each prompt. The generated tokens are verified by LLM and the numbers of correct tokens are recorded. We denote the number of correct tokens generated for the $i$-th prompt as $l_i$. For an $s \leq 80$, we can use $l_i$-s to approximate $l(s)$, the average number of correct tokens in the first $s$ tokens, as follows:
\begin{equation}
l(s) \approx \frac{1}{n} \sum_{i=1}^{n} \min \left\{ l_i, s \right\}
\end{equation}
The measured $l(s)$ curve is shown in the blue curve in Figure~\ref{fig:expected-len}. We can use a power function \begin{equation}
l(s) \approx c\cdot s^\gamma
\end{equation} to approximate it. In this case, the approximation function is $0.9s^{0.548}$, which is plotted as the orange curve.




This is also consistent with the intuition that $l(s)$ is a non-decreasing and sub-linear function ($\gamma$ is always smaller than 1).
This is because we can decompose $l(s)$ as following:
\begin{equation}
    l(s)=\sum_{i=1}^{s} p(\bigwedge_{j=1}^i E_i)
\end{equation}
where $p$ means probability and $E_j$ is the random event that the $j$-th token is correct. Every term in the summation is non-negative. As $s$ increases, it is harder to make correct speculation, so the probability $p$ will decrease rapidly. 


\noindent\textbf{LLM Runtime for Each Speculation Step $t_L(b, s)$} 

\begin{figure}
    \centering
    \includegraphics[width=0.8\linewidth]{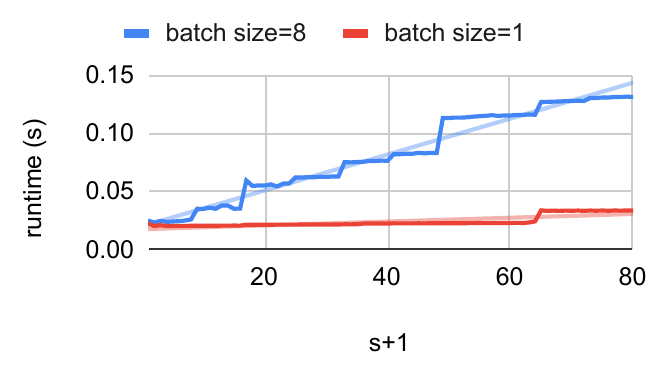}
    \caption{Execution time $t_L(B, s)$ achieved when varying the speculation length $s$ for different batch sizes $B$-s, which can be approximated by a linear function. 
    }
    \label{fig:seqlen-time}
\end{figure}

The runtime of one verification step of LLM $t_L(b, s)$ depends on the hardware and model implementations. To illustrate how it scales with $s$, 
we measured the runtime of OPT-6.7B on an RTX 3090 GPU, and the result is shown in Figure~\ref{fig:seqlen-time}. The runtime curves are approximately step functions. The first jump of the curve of $b=1$ is when $s=64$, while that of $B=8$ jumps up much earlier when $s=8$. For simplicity, we approximate $t_L(b, s)$ as linear functions with different slopes $t_L(b, s) = \alpha_b s + \beta$.  

\subsubsection{SSM Runtime $T_S$} 

In practice, we usually also need to consider the second term, i.e. the runtime of the small model. For a fixed $B$, $t_S(B,2)$ is a constant, and the constant is also increasing with respect to $B$. 

\subsubsection{Optimal Speculation Length $s_{opt}$}

Based on the aforementioned analysis, the total runtime can be written as
\begin{align}
t &= \frac{N \cdot t_L(b, s)}{l(s) + 1} + \frac{N \cdot s \cdot t_S(b, 1)}{l(s) + 1} \\
&\approx N \frac{\alpha_b \cdot s + \beta + t_S(b, 1)\cdot s}{c\cdot s^\gamma + 1} 
\end{align}
Since $t_S(b,1)$ is a constant regarding $s$, and is also increasing with $b$ as $\alpha_b$, we merge it with $\alpha_b$ for simplicity.

The minimum point of this function $s_{opt}$ is where the derivative of $t$ is zero, i.e. $\frac{\partial t}{\partial s} = 0$. And we have
\begin{equation}
\frac{\partial t}{\partial s} = N\frac{\alpha_b(c\cdot s^\gamma + 1) - (\alpha_b \cdot s + \beta) \cdot c \gamma s^{\gamma - 1}}{(c\cdot s^\gamma + 1)^2}
\end{equation}
It is zero when the numerator is zero, which can be written as
\begin{equation}(1-\gamma)\alpha_b \cdot c s^\gamma - c\cdot \beta \cdot \gamma s^{\gamma - 1} + \alpha_b
\end{equation}
Let $K=(1-\gamma)\cdot c$, $L=c\cdot \beta \cdot \gamma$, we can rewrite the expression above in the following form:
\begin{equation}
\delta = K\alpha_b \cdot s^\gamma - Ls^{\gamma - 1} + \alpha_b
\end{equation}
The optimal speculation length $s_{opt}$ needs to satisfy $\delta=0$. 
\begin{equation}
 K\alpha_b \cdot s^\gamma - Ls^{\gamma - 1} + \alpha_b = 0
\end{equation}
Since $K \cdot s^\gamma > 0$, $Ls^{\gamma-1}$ is a constant regarding $b$, and $\alpha_b$ is increasing with $b$, $\delta$ is monotonically increasing with respect to $b$. Similarly, because $K\cdot \alpha_b > 0$ and $\gamma - 1 < 0$, $\delta$ is also increasing with $s$. Therefore, for a larger $b$, $s_{opt}$ needs to be smaller in order to keep the $\delta$ as zero, and vice versa, i.e., for a smaller b, $s_{opt}$ needs to be larger.



\section{Adaptive Speculative Decoding: Our Approach}


In real-world serving scenarios, the traffic might be time-varying, leading to varying batch sizes. Based on the observation and analysis that the optimal speculation length depends on the batch size, we propose a new adaptive speculation mechanism to choose the optimal speculation length under different scenarios.

The key point is to find the optimal speculation step $s_{opt}$ for each different batch size $b$. While it is hard to find a closed-form solution for it, we can empirically measure the and do a grid search to find $s_{opt}$. We divide the process into two stages, namely profiling and execution. In the profiling stage, we use a small sample of the dataset (similar to the training set in machine learning) to measure the latency under different batch sizes using different speculation lengths. 
Using the profiling result, we can build a look-up table (LUT) that stores the optimal speculation length for each batch size. During the execution, after a batch of requests arrives, we can find the optimal speculation length using the LUT, and configure the LLM execution, accordingly. 

The overhead of profiling is negligible in LLM serving scenarios for two key reasons. First, we enable a very small search space. The optimal speculation length is usually small (less than ten). While profiling all possible batch sizes can be prohibitive. Instead, we sample only a subset of them. We profile batch sizes which are powers of two (1, 2, 4, 8, 16, ...). For batch sizes that are not profiled, we choose the smaller speculation length of the nearest two profiled batch sizes. 
Second, the profiling overhead can be amortized, because LLM serving is usually a long-running serving task. A service can be online for weeks or months serving millions or even billions of user requests, while profiling only takes several minutes and it is only done once: before launching the LLM inference process in the server.


\vspace{15pt}
\section{Evaluation}

We evaluate the performance benefits of our proposed adaptive batched speculative mechanism using the methodology presented in Section~\ref{sec:methodology} in two aspects: (i) quantifying a uniform traffic, where user requests are served by LLM at a fixed batch size (Section~\ref{sec:eval-fixed}), and (ii) quantifying a dynamic traffic, where user requests are served by LLM at a time variable batch size (Section~\ref{sec:eval-dynamic}).

\subsection{Methodology}\label{sec:methodology}

\paragraph{Dataset.} We use the \textit{Chatbot Instruction Prompts} dataset~\cite{dataset} from Huggingface. We only use the prompt field of it as the input.

\paragraph{Hardware and Software.} We use NVIDIA RTX 3090 as our benchmark platform. We use Ubuntu 22.04, CUDA 11.8, and PyTorch 2.0.1.

\paragraph{Models.} We use OPT-6.7B~\cite{opt} as our large language model (LLM) and OPT-125M as the small speculative model (SSM).

\subsection{Uniform Traffic}\label{sec:eval-fixed}
\begin{figure}[h]
    \centering
    \includegraphics[width=0.48\textwidth]{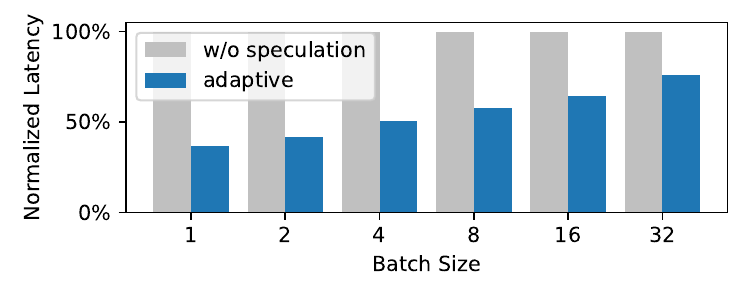}
    \caption{Normalized latency with fixed batch sizes. }
    \label{fig:static-latency}
\end{figure}

In this section, we assume that user requests are served by LLM with a fixed batch size. We compare our approach, i.e., adaptive speculation decoding, that integrates using the optimal speculation length, over the baseline LLM serving without speculation decoding. We sample 1000 prompts from the dataset and group them into batches. For each prompt, we generate 128 new tokens. Figure~\ref{fig:static-latency} presents the end-to-end time of finishing all prompts, including the tokenizer, by normalizing the results over the baseline without speculation decoding. 
We did a grid search to find the optimal speculation length.

We make two observations. First, adaptive speculative decoding provides higher performance benefits for smaller batch sizes. For example, when the batch size is 1, adaptive speculating decoding achieves a 2.73$\times$ speedup (63\% latency reduction), while when the batch size is 32, the speedup is 1.31$\times$. This is because when batch size is small GPUs are highly underutilized, and it is leveraged by speculative decoding. When batch size is large, GPUs are well utilized, so the room for performance improvement is limited. Second, across all batch sizes, adaptive speculative decoding provides a high speedup by on average of 1.94$\times$ over LLM serving with no speculative decoding.


\subsection{Dynamic Traffic}\label{sec:eval-dynamic}
In this section, we measure the effectiveness of adaptive speculative decoding in a real-world scenario, where the traffic highly varies over time. We create a server-client setting, and we artificially vary the number of requests sent to the LLM server per second. 

\begin{figure}[h]
    \centering
    \begin{subfigure}[b]{0.23\textwidth}
        \centering
        \includegraphics[width=\textwidth]{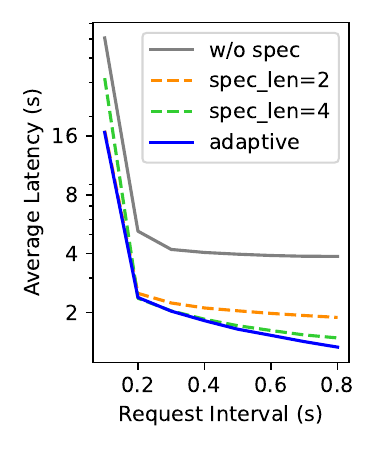}
        \caption{cv=0.5}
        \label{fig:cn.5}
    \end{subfigure}
    \hfill
    \begin{subfigure}[b]{0.23\textwidth}
        \centering
        \includegraphics[width=\textwidth]{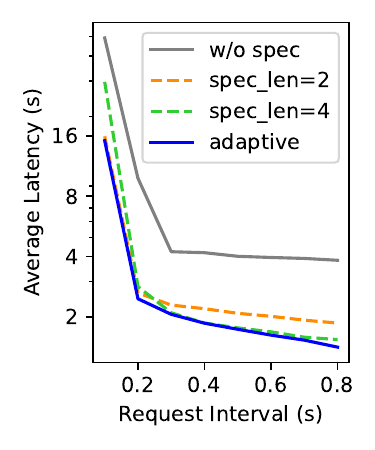}
        \caption{cv=1}
        \label{fig:cv1}
    \end{subfigure}
    \\
    \begin{subfigure}[b]{0.23\textwidth}
        \centering
        \includegraphics[width=\textwidth]{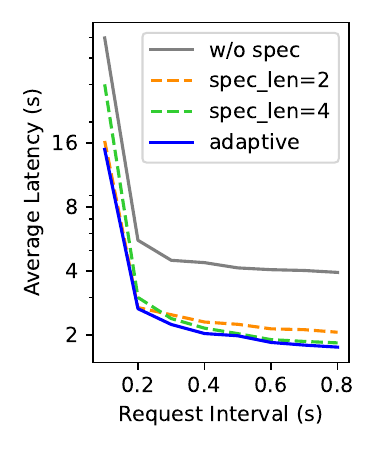}
        \caption{cv=2}
        \label{fig:cv2}
    \end{subfigure} \hfill
    \begin{subfigure}[b]{0.23\textwidth}
        \centering
        \includegraphics[width=\textwidth]{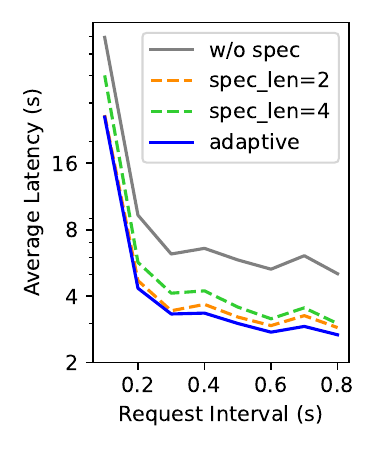}
        \caption{cv=5}
        \label{fig:cv5}
    \end{subfigure}
    \caption{Average latency with various traffic scenarios, i.e., varying request intervals and coefficient of variation (CV).}
    \label{fig:latency}
\end{figure}

\begin{figure*}[ht]
    \centering
    \begin{subfigure}[t]{0.75\textwidth}
        \centering
        \includegraphics[width=\textwidth]{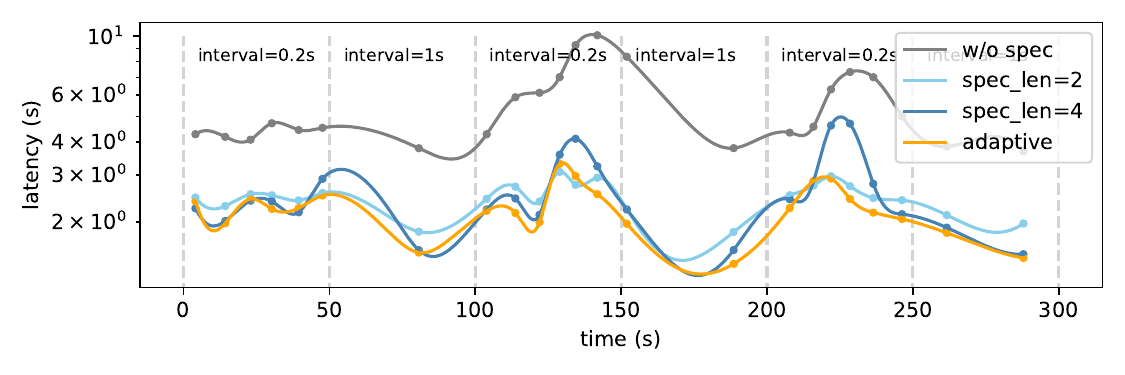}
    \end{subfigure}\hfill
    \begin{subfigure}[t]{0.25\textwidth}
        \centering
        \includegraphics[width=\textwidth]{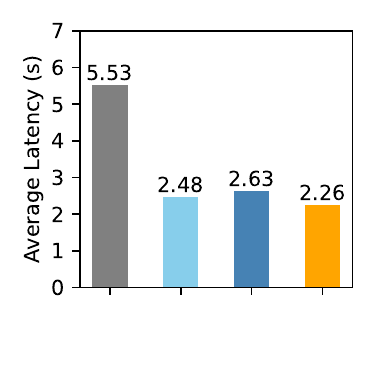}
    \end{subfigure}

    \caption{Timeline of four speculation strategies. Each point represents a group of 40 requests. The X-axis is the time stamp of the first request in a group. The Y-axis is the average latency of all requests in a group of requests. There is no optimal speculation length for all scenarios. When the request traffic is intense (e.g. 100-150, 200-250), a smaller speculation length like 2 is preferred. When the request traffic is sparse (e.g. 150-200, 250-300), a larger speculation length like 4 is preferred. Nevertheless, adaptive speculation decoding can always achieve performance on par with or better than both fixed speculation length schemes.}
    \label{fig:timeline}
\end{figure*}

\paragraph{Client Configuration.} We launch a client process that samples a sequence of requests from the dataset and sends them to the server at a fixed interval time. The interval time is sampled from a Gamma distribution, and the parameters of the Gamma distribution allow us to control the average interval between requests as well as the coefficient of variation (CV). This way we can control the overall volume of traffic by varying the average interval, and the variation and blast of the traffic by varying the CV.

\paragraph{Server Configuration.} We launch a server process and wrap the LLM inference as a service that receives requests from a message queue and responds to the generated tokens via another message queue. If there is more than one request in the queue, they will be merged as one batched request (up to a maximal batch size of 16 due to the memory constraint).

\paragraph{Comparison Points.} We compare our proposed adaptive speculative decoding scheme with three schemes: 1) Batched LLM inference without speculative decoding, 2) Batched LLM inference with speculative decoding of fixed speculation length equal to 2, and 3) Batched LLM inference with speculative decoding of fixed speculation length. For fixed speculative decoding schemes, we select speculation lengths of 2 and 4, since they achieve better performance. We do not evaluate the speculative decoding scheme without batching, because it performs significantly worse than our remaining evaluated schemes.

\paragraph{Experiments.} We create dynamic request traffic on the client side and evaluate various volume and blast traffic configurations using four different values for CV (0.5, 1, 2, and 5) and eight different values for the interval time from 0.1s up to 0.8s with a step of 0.1s. For each setting, we generate only one sequence of requests, which is used to evaluate all comparison points. Each 'sequence' contains 1000 prompts, and there are no overlaps between the dataset used in the profiling step of our proposed scheme and the dataset used in our dynamic traffic evaluation. We record the time $t_a$ at which each request is sent by the client to the server and the time $t_b$ at which the server finishes serving the request and measure the latency as the difference $t_b - t_a$. Note that if the server is slow, requests cannot be processed in time and have to wait in a queue, thus the waiting time is also accounted for in the latency. Figure~\ref{fig:latency} presents the average latency across all requests achieved by all comparison points using different configurations for dynamic traffic generation.


We draw two findings. First, adaptive speculative decoding can provide high latency benefits in all various scenarios, including intense (low request interval) and sparse (high request interval) traffic, as well as uniform (low CV) and skewed (high CV) traffic scenarios. On average, adaptive speculative decoding achieves a speedup of 2.3$\times$ over LLM serving with no speculative decoding. Second, we observe that the best-performing speculation length varies across different traffic scenarios, i.e., there is no speculation length value that provides the lowest latency across all traffic scenarios. For example, when having CV=2 and a request interval equal to 0.1, a speculation length of 2 achieves the best latency, while when having CV=2 and and request interval equal to 0.6, a speculation length of 4 achieves the best latency. We observe that our proposed scheme effectively adapts the speculation length to the characteristics of the current traffic and achieves the best performance, outperforming both fixed speculation decoding schemes. Especially when the variation of the traffic is large (e.g. when CV is equal to 5), because the batch size might vary with the traffic volume, adaptive speculative decoding provides an additional $1.07\times$ average speedup over the better one of the two fixed speculation length.
The speedup over fixed speculation length can be up to 1.15$\times$. 




Figure~\ref{fig:timeline}
presents the timeline on latency achieved by all schemes during runtime under time-varying traffic scenarios. In this experiment, the client alternates between two types of traffic: \emph{intense} with a request interval of 0.2 seconds and \emph{sparse} with a request interval of 1 second. The client will switch from one type of traffic to the other every 50 seconds. We fixed the CV as 1 in this experiment.

We find that adaptive speculation decoding effectively adapts to best-performing configuration. A speculation length of 4 performs better for sparse request traffic because the batch size is usually small, and a speculation length of 2 is more suitable for intense traffic. For example, from the 150th to 200th second, the traffic is sparse, therefore the speculation length of 4 performs better than the speculation length of 2. The situation is opposite from the 200th to 250th seconds. Nevertheless, adaptive speculation decoding can always match the performance of the better one of the two static baselines, or even perform better. Overall, adaptive improves latency over speculation lengths of 2 and 4 by 9\% and 14\%, respectively, on average. We conclude that our approach provides significant performance benefits by dynamically tuning itself based on the characteristics of the current traffic and leveraging the best of both speculative decoding and batching optimizations. 


\section{Conclusion}

LLMs like GPT achieve significant success nowadays. However, LLM execution is inherently sequential, thus incurring low hardware utilization on GPUs. Batching and speculative decoding are two techniques to improve the GPU hardware utilization in LLM inference. To study the synergy between them, we carry out a comprehensive study, and observe that the optimal speculation length depends on the batch size as well as the characteristics of LLM model and GPU architecture. We build a quantitative model to explain this key observation, and based on our analysis, we propose an efficient adaptive speculative decoding strategy that selects the optimal speculation length for different batch sizes. Our evaluations show that our proposed method achieves equal or better performance than fixed speculation decoding schemes, and we hope our work will enable further research on accelerating LLM serving.

\bibliography{ref}

\begin{thebibliography}{22}
\providecommand{\natexlab}[1]{#1}
\providecommand{\url}[1]{\texttt{#1}}
\expandafter\ifx\csname urlstyle\endcsname\relax
  \providecommand{\doi}[1]{doi: #1}\else
  \providecommand{\doi}{doi: \begingroup \urlstyle{rm}\Url}\fi

\bibitem[Brohan et~al.(2023)Brohan, Brown, Carbajal, Chebotar, Chen, Choromanski, Ding, Driess, Dubey, Finn, et~al.]{brohan2023rt}
Brohan, A., Brown, N., Carbajal, J., Chebotar, Y., Chen, X., Choromanski, K., Ding, T., Driess, D., Dubey, A., Finn, C., et~al.
\newblock Rt-2: Vision-language-action models transfer web knowledge to robotic control.
\newblock \emph{arXiv preprint arXiv:2307.15818}, 2023.

\bibitem[Brown et~al.(2020)Brown, Mann, Ryder, Subbiah, Kaplan, Dhariwal, Neelakantan, Shyam, Sastry, Askell, et~al.]{gpt3}
Brown, T., Mann, B., Ryder, N., Subbiah, M., Kaplan, J.~D., Dhariwal, P., Neelakantan, A., Shyam, P., Sastry, G., Askell, A., et~al.
\newblock Language models are few-shot learners.
\newblock \emph{Advances in neural information processing systems}, 33:\penalty0 1877--1901, 2020.

\bibitem[Chen et~al.(2023)Chen, Borgeaud, Irving, Lespiau, Sifre, and Jumper]{specsample}
Chen, C., Borgeaud, S., Irving, G., Lespiau, J.-B., Sifre, L., and Jumper, J.
\newblock Accelerating large language model decoding with speculative sampling.
\newblock \emph{arXiv preprint arXiv:2302.01318}, 2023.

\bibitem[Chen et~al.(2021)Chen, Tworek, Jun, Yuan, Pinto, Kaplan, Edwards, Burda, Joseph, Brockman, et~al.]{llm4code}
Chen, M., Tworek, J., Jun, H., Yuan, Q., Pinto, H. P. d.~O., Kaplan, J., Edwards, H., Burda, Y., Joseph, N., Brockman, G., et~al.
\newblock Evaluating large language models trained on code.
\newblock \emph{arXiv preprint arXiv:2107.03374}, 2021.

\bibitem[Crankshaw et~al.(2017)Crankshaw, Wang, Zhou, Franklin, Gonzalez, and Stoica]{crankshaw2017clipper}
Crankshaw, D., Wang, X., Zhou, G., Franklin, M.~J., Gonzalez, J.~E., and Stoica, I.
\newblock Clipper: A $\{$Low-Latency$\}$ online prediction serving system.
\newblock In \emph{14th USENIX Symposium on Networked Systems Design and Implementation (NSDI 17)}, pp.\  613--627, 2017.

\bibitem[Fang et~al.(2021)Fang, Yu, Zhao, and Zhou]{fang2021turbotransformers}
Fang, J., Yu, Y., Zhao, C., and Zhou, J.
\newblock Turbotransformers: an efficient gpu serving system for transformer models.
\newblock In \emph{Proceedings of the 26th ACM SIGPLAN Symposium on Principles and Practice of Parallel Programming}, pp.\  389--402, 2021.

\bibitem[Gao et~al.(2018)Gao, Yu, Wu, and Li]{gao2018low}
Gao, P., Yu, L., Wu, Y., and Li, J.
\newblock Low latency rnn inference with cellular batching.
\newblock In \emph{Proceedings of the Thirteenth EuroSys Conference}, pp.\  1--15, 2018.

\bibitem[Ivanov et~al.(2021)Ivanov, Dryden, Ben-Nun, Li, and Hoefler]{ivanov2021data}
Ivanov, A., Dryden, N., Ben-Nun, T., Li, S., and Hoefler, T.
\newblock Data movement is all you need: A case study on optimizing transformers.
\newblock \emph{Proceedings of Machine Learning and Systems}, 3:\penalty0 711--732, 2021.

\bibitem[Lee et~al.(2018)Lee, Scolari, Chun, Santambrogio, Weimer, and Interlandi]{lee2018pretzel}
Lee, Y., Scolari, A., Chun, B.-G., Santambrogio, M.~D., Weimer, M., and Interlandi, M.
\newblock $\{$PRETZEL$\}$: Opening the black box of machine learning prediction serving systems.
\newblock In \emph{13th USENIX Symposium on Operating Systems Design and Implementation (OSDI 18)}, pp.\  611--626, 2018.

\bibitem[Leviathan et~al.(2023)Leviathan, Kalman, and Matias]{fastinferspec}
Leviathan, Y., Kalman, M., and Matias, Y.
\newblock Fast inference from transformers via speculative decoding.
\newblock In \emph{International Conference on Machine Learning}, pp.\  19274--19286. PMLR, 2023.

\bibitem[Liu et~al.(2023)Liu, Hu, Bailis, Stoica, Deng, Cheung, and Zhang]{liu2023online}
Liu, X., Hu, L., Bailis, P., Stoica, I., Deng, Z., Cheung, A., and Zhang, H.
\newblock Online speculative decoding.
\newblock \emph{arXiv preprint arXiv:2310.07177}, 2023.

\bibitem[Miao et~al.(2023)Miao, Oliaro, Zhang, Cheng, Wang, Wong, Chen, Arfeen, Abhyankar, and Jia]{specinfer}
Miao, X., Oliaro, G., Zhang, Z., Cheng, X., Wang, Z., Wong, R. Y.~Y., Chen, Z., Arfeen, D., Abhyankar, R., and Jia, Z.
\newblock Specinfer: Accelerating generative llm serving with speculative inference and token tree verification.
\newblock \emph{arXiv preprint arXiv:2305.09781}, 2023.

\bibitem[Olston et~al.(2017)Olston, Fiedel, Gorovoy, Harmsen, Lao, Li, Rajashekhar, Ramesh, and Soyke]{olston2017tensorflow}
Olston, C., Fiedel, N., Gorovoy, K., Harmsen, J., Lao, L., Li, F., Rajashekhar, V., Ramesh, S., and Soyke, J.
\newblock Tensorflow-serving: Flexible, high-performance ml serving.
\newblock \emph{arXiv preprint arXiv:1712.06139}, 2017.

\bibitem[OpenAI(2023)]{gpt4}
OpenAI.
\newblock Gpt-4 technical report, 2023.

\bibitem[Palla(2023)]{dataset}
Palla, A.
\newblock Chatbot instruction prompts datasets.
\newblock \url{https://huggingface.co/datasets/alespalla/chatbot_instruction_prompts}, 2023.

\bibitem[Shen et~al.(2019)Shen, Chen, Jin, Zhao, Kong, Philipose, Krishnamurthy, and Sundaram]{shen2019nexus}
Shen, H., Chen, L., Jin, Y., Zhao, L., Kong, B., Philipose, M., Krishnamurthy, A., and Sundaram, R.
\newblock Nexus: A gpu cluster engine for accelerating dnn-based video analysis.
\newblock In \emph{Proceedings of the 27th ACM Symposium on Operating Systems Principles}, pp.\  322--337, 2019.

\bibitem[Spector \& Re(2023)Spector and Re]{spector2023accelerating}
Spector, B. and Re, C.
\newblock Accelerating llm inference with staged speculative decoding.
\newblock \emph{arXiv preprint arXiv:2308.04623}, 2023.

\bibitem[Thirunavukarasu et~al.(2023)Thirunavukarasu, Ting, Elangovan, Gutierrez, Tan, and Ting]{thirunavukarasu2023large}
Thirunavukarasu, A.~J., Ting, D. S.~J., Elangovan, K., Gutierrez, L., Tan, T.~F., and Ting, D. S.~W.
\newblock Large language models in medicine.
\newblock \emph{Nature medicine}, 29\penalty0 (8):\penalty0 1930--1940, 2023.

\bibitem[Touvron et~al.(2023)Touvron, Lavril, Izacard, Martinet, Lachaux, Lacroix, Rozière, Goyal, Hambro, Azhar, Rodriguez, Joulin, Grave, and Lample]{llama}
Touvron, H., Lavril, T., Izacard, G., Martinet, X., Lachaux, M.-A., Lacroix, T., Rozière, B., Goyal, N., Hambro, E., Azhar, F., Rodriguez, A., Joulin, A., Grave, E., and Lample, G.
\newblock Llama: Open and efficient foundation language models, 2023.

\bibitem[Wolf et~al.(2019)Wolf, Debut, Sanh, Chaumond, Delangue, Moi, Cistac, Rault, Louf, Funtowicz, et~al.]{huggingface}
Wolf, T., Debut, L., Sanh, V., Chaumond, J., Delangue, C., Moi, A., Cistac, P., Rault, T., Louf, R., Funtowicz, M., et~al.
\newblock Huggingface's transformers: State-of-the-art natural language processing.
\newblock \emph{arXiv preprint arXiv:1910.03771}, 2019.

\bibitem[Yu et~al.(2022)Yu, Jeong, Kim, Kim, and Chun]{orca}
Yu, G.-I., Jeong, J.~S., Kim, G.-W., Kim, S., and Chun, B.-G.
\newblock Orca: A distributed serving system for $\{$Transformer-Based$\}$ generative models.
\newblock In \emph{16th USENIX Symposium on Operating Systems Design and Implementation (OSDI 22)}, pp.\  521--538, 2022.

\bibitem[Zhang et~al.(2022)Zhang, Roller, Goyal, Artetxe, Chen, Chen, Dewan, Diab, Li, Lin, et~al.]{opt}
Zhang, S., Roller, S., Goyal, N., Artetxe, M., Chen, M., Chen, S., Dewan, C., Diab, M., Li, X., Lin, X.~V., et~al.
\newblock Opt: Open pre-trained transformer language models.
\newblock \emph{arXiv preprint arXiv:2205.01068}, 2022.

\end{thebibliography}
\bibliographystyle{icml2023}

\end{document}